\documentclass[conference,english,11pt]{article}
\usepackage{url}
\usepackage[T1]{fontenc}
\usepackage[latin1]{inputenc}
\usepackage[pdftex]{graphicx}
\usepackage{subfig}
\usepackage{amssymb}
\usepackage{babel}
\usepackage{amsmath, amsthm, amssymb, amsfonts, mathtools}
\usepackage{color}
\usepackage{longtable}
\usepackage{booktabs}
\usepackage{pgfgantt}
\usepackage{grfext}
\usepackage{acronym}
\usepackage{units}
\usepackage{tabularx}
\usepackage{algpseudocode}
\usepackage{algorithmicx}
\usepackage{authblk}
\usepackage[square,sort]{natbib}
\graphicspath{{./img/}}
\acrodef{AC}[AC]{Arrenhius \& Current}
\acrodef{AER}[AER]{Address Event Representation}
\acrodef{AEX}[AEX]{AER EXtension board}
\acrodef{AMDA}[AMDA]{``AER Motherboard with D/A converters''}
\acrodef{API}[API]{Application Programming Interface}
\acrodef{BM}[BM]{Boltzmann Machine}
\acrodef{BP}[BP]{Back Propagation}
\acrodef{CAVIAR}[CAVIAR]{Convolution AER Vision Architecture for Real-Time}
\acrodef{CCN}[CCN]{Cooperative and Competitive Network}
\acrodef{CD}[CD]{Contrastive Divergence}
\acrodef{eCD}[eCD]{event-driven Contrastive Divergence}
\acrodef{CMOS}[CMOS]{Complementary Metal--Oxide--Semiconductor}
\acrodef{COTS}[COTS]{Commercial Off-The-Shelf}
\acrodef{CPU}[CPU]{Central Processing Unit}
\acrodef{CV}[CV]{Coefficient of Variation}
\acrodef{CV}[CV]{Coefficient of Variation}
\acrodef{DAC}[DAC]{Digital--to--Analog}
\acrodef{DBN}[DBN]{Deep Belief Network}
\acrodef{DFA}[DFA]{Deterministic Finite Automaton}
\acrodef{DFA}[DFA]{Deterministic Finite Automaton}
\acrodef{divmod3}[DIVMOD3]{divisibility of a number by 3}
\acrodef{DPE}[DPE]{Dynamic Parameter Estimation}
\acrodef{DPI}[DPI]{Differential-Pair Integrator}
\acrodef{DSP}[DSP]{Digital Signal Processor}
\acrodef{DVS}[DVS]{Dynamic Vision Sensor}
\acrodef{EDVAC}[EDVAC]{Electronic Discrete Variable Automatic Computer}
\acrodef{EIF}[EI\&F]{Exponential Integrate \& Fire}
\acrodef{EIN}[EIN]{Excitatory--Inhibitory Network}
\acrodef{EPSC}[EPSC]{Excitatory Post-Synaptic Current}
\acrodef{EPSP}[EPSP]{Excitatory Post--Synaptic Potential}
\acrodef{FPGA}[FPGA]{Field Programmable Gate Array}
\acrodef{FSM}[FSM]{Finite State Machine}
\acrodef{GPU}[GPU]{Graphical Processing Unit}
\acrodef{HAL}[HAL]{Hardware Abstraction Layer}
\acrodef{HH}[H\&H]{Hodgkin \& Huxley}
\acrodef{HMM}[HMM]{Hidden Markov Model}
\acrodef{HW}[HW]{Hardware}
\acrodef{hWTA}[hWTA]{Hard Winner--Take--All}
\acrodef{IF2DWTA}[IF2DWTA]{Integrate \& Fire 2--Dimensional WTA}
\acrodef{IF}[I\&F]{Integrate \& Fire}
\acrodef{IFSLWTA}[IFSLWTA]{Integrate \& Fire Stop Learning WTA}
\acrodef{INCF}[INCF]{International Neuroinformatics Coordinating Facility}
\acrodef{INI}[INI]{Institute of Neuroinformatics}
\acrodef{IO}[IO]{Input-Output}
\acrodef{IPSC}[IPSC]{Inhibitory Post-Synaptic Current}
\acrodef{ISI}[ISI]{Inter--Spike Interval}
\acrodef{JFLAP}[JFLAP]{Java - Formal Languages and Automata Package}
\acrodef{LIF}[LI\&F]{Linear Integrate \& Fire}
\acrodef{LSM}[LSM]{Liquid State Machine}
\acrodef{LTD}[LTD]{Long-Term Depression}
\acrodef{LTI}[LTI]{Linear Time-Invariant}
\acrodef{LTP}[LTP]{Long-Term Potentiation}
\acrodef{LTU}[LTU]{Linear Threshold Unit}
\acrodef{MCMC}{Markov Chain Monte Carlo}
\acrodef{NHML}[NHML]{Neuromorphic Hardware Mark-up Language}
\acrodef{NMDA}[NMDA]{NMDA}
\acrodef{NME}[NE]{Neuromorphic Engineering}
\acrodef{PCB}[PCB]{Printed Circuit Board}
\acrodef{PRC}[PRC]{Phase Response Curve}
\acrodef{PSC}[PSC]{Post-Synaptic Current}
\acrodef{PSP}[PSP]{Post--Synaptic Potential}
\acrodef{RI}[KL]{Kullback-Leibler}
\acrodef{RRAM}[RRAM]{Resistive Random-Access Memory}
\acrodef{RBM}[RBM]{Restricted Boltzmann Machine}
\acrodef{ROC}[ROC]{Receiver Operator Characteristic}
\acrodef{SAC}[SAC]{Selective Attention Chip}
\acrodef{SSM}[S2M]{Synaptic Sampling Machine}
\acrodef{dSSM}[S2M]{Synaptic Sampling Machine}
\acrodef{S3M}[spiking S2M]{Spiking S2M}
\acrodef{SCD}[SCD]{Spike-Based Contrastive Divergence}
\acrodef{SCX}[SCX]{Silicon CorteX}
\acrodef{STDP}[STDP]{Spike Timing Dependent Plasticity}
\acrodef{SW}[SW]{Software}
\acrodef{sWTA}[SWTA]{Soft Winner--Take--All}
\acrodef{VHDL}[VHDL]{VHSIC Hardware Description Language}
\acrodef{VLSI}[VLSI]{Very  Large  Scale  Integration}
\acrodef{WTA}[WTA]{Winner--Take--All}
\acrodef{XML}[XML]{eXtensible Mark-up Language}

\usepackage[margin=.75in]{geometry}
\normalsize
\newcommand{\largefigwidth}{1.0\textwidth}
\newcommand{\figwidth}{.46\textwidth}
\renewcommand{\refeq}[1]{{Eq.~(\ref{#1})}}

\newcommand{\reffig}[1]{{Fig.~\ref{#1}}}
\newcommand{\reftab}[1]{{Tab.~\ref{#1}}}

\newcommand{\vectwo}[2]{\left(\begin{matrix}#1 \\ #2 \end{matrix}\right)}
\renewcommand{\cite}[1]{{\color{black!40}\citep{#1}}}

\begin{document}
%
\title{Neuromorphic Deep Learning Machines}

\author[1]{Emre Neftci}
\author[3]{Charles Augustine}
\author[3]{Somnath Paul}
\author[1]{Georgios Detorakis}
\affil[1]{Department of Cognitive Sciences, UC Irvine, Irvine, CA, USA\\}
\affil[3]{Circuit Research Lab, Intel Corporation, Hilsboro, OR, USA\\}

\maketitle

\begin{abstract}
An ongoing challenge in neuromorphic computing is to devise general and computationally efficient models of inference and learning which are compatible with the spatial and temporal constraints of the brain.
One increasingly popular and successful approach is to take inspiration from inference and learning algorithms used in deep neural networks.
However, the workhorse of deep learning, the gradient descent \ac{BP} rule, often relies on the immediate availability of network-wide information stored with high-precision memory, and precise operations that are difficult to realize in neuromorphic hardware. 
Remarkably, recent work showed that exact backpropagated weights are not essential for learning deep representations. Random \ac{BP} replaces feedback weights with random ones and encourages the network to adjust its feed-forward weights to learn pseudo-inverses of the (random) feedback weights.
Building on these results, we demonstrate an event-driven random \ac{BP} (eRBP) rule that uses an error-modulated synaptic plasticity for learning deep representations in neuromorphic computing hardware.
The rule requires only one addition and two comparisons for each synaptic weight using a two-compartment leaky \ac{IF} neuron, making it very suitable for implementation in digital or mixed-signal neuromorphic hardware.
Our results show that using eRBP, deep representations are rapidly learned, achieving nearly identical classification accuracies compared to artificial neural network simulations on GPUs, while being robust to neural and synaptic state quantizations during learning.
\end{abstract}


\section{Introduction}
Biological neurons and synapses can provide the blueprint for inference and learning machines that are potentially thousandfold more energy efficient than mainstream computers.
However, the breadth of application and scale of present-day neuromorphic hardware remains limited, mainly due to a lack of general and efficient inference and learning algorithms compliant with the spatial and temporal constraints of the brain.\\
Machine learning and deep learning are well poised for solving a broad set of applications using neuromorphic hardware, thanks to their general-purpose, modular, and fault-tolerant nature \cite{Esser_etal16,Neftci_etal16,Lee_etal16}.
One outstanding question is whether the learning phase in deep neural networks can be efficiently carried out in neuromorphic hardware.
Performing learning on-the-fly in less controlled environments where no prior, representative dataset exists confers more fine-grained context awareness to behaving cognitive agents.
However, deep learning usually relies on the immediate availability of network-wide information stored with high-precision memory.
In digital computers, the access to this information funnels through the von Neumann bottleneck, which dictates the fundamental limits of the computing substrate.    
Distributing computations along multiple cores (such as in GPUs) is a popular solution to mitigate this problem, but even there the scalability of BP is often limited by its memory-intensive operations \cite{Zhu_etal16}.

The implementation of gradient \ac{BP} on a neural substrate is even more challenging \cite{Baldi_etal16,Lee_etal16,Grossberg87} because it requires 1) using synaptic weights that are identical with forward passes (symmetric weights requirements, also known as the weight transport problem), 2) carrying out the operations involved in BP including multiplications with derivatives and activation functions, 3) propagating error signals with high precision, 4) alternating between forward and backward passes, 5) changing the sign of synaptic weights, and 6) availability of targets (labels).
The essence of these challenges is that it requires precise linear and non-linear computations, and more importantly because gradient \ac{BP} requires information that is not local to the computational building blocks in a neural substrate, meaning that special communcation channels must be provisioned. 
Whether a given operation is local or not depends on the physical substrate that carries out the computations.
For example, while symmetric weights in neural networks are compatible with von Neumann architectures (and even desirable since weights in both directions are shared), the same is not true in a distributed system such as the brain: elementary computing units do not have bidirectional connections with the same weight in each direction.
Since neuromorphic implementations generally assume dynamics closely related to the those in the brain, requirements (1-4) above also hinder efficient implementations of \ac{BP} in neuromorphic hardware.

Although previous work \cite{OConnor_Welling16,Lee_etal16,Lillicrap_etal16} overcomes some of the fundamental difficulties of gradient \ac{BP} listed above in spiking networks, here we tackle all of the key difficulties using event-driven random \ac{BP} (eRBP), a learning rule for deep spiking neural networks achieving classification accuracies that are similar to those obtained in artificial neural networks, potentially running on a fraction of the energy budget with dedicated neuromorphic hardware.

eRBP builds on the recent advances in approximate forms of the gradient BP rule \cite{Lee_etal14,Liao_etal15,Lillicrap_etal16,Baldi_etal16} for training spiking neurons of the type used in neuromorphic hardware to perform supervised learning.
These approximations solve the non-locality problem by replacing BP weights with random ones, leading to remarkably little loss in classification performance on benchmark tasks \cite{Lillicrap_etal16,Baldi_etal16} (requirement 1 above). 
Although a general theoretical understanding of random BP (RBP) is still lacking, extended simulations and analyses of linear networks show that, during learning, the network adjusts its feed-forward weights to learn an approximation of the pseudo-inverse of the (random) feedback weights, which is equally good in communicating gradients.
eRBP is an asynchronous (event-driven) adaptation of random \ac{BP} that can be tightly embedded with the dynamics of dual compartment \ac{IF} neurons that costs one addition and two comparisons per synaptic weight update.
Extended experimentations show that the spiking nature of neuromorphic hardware and the lack of general linear and non-linear computations at the neuron does not prevent accurate learning on classification tasks (requirement 2, 3), and operates continuously and asynchronously without alternation of forward or backward passes (requirement 4).
%
Additional experimental evidence shows that  eRBP is robust to fixed width representations with limited neural and synaptic state precision, making it suitable for dedicated digital hardware. 
The success of eRBP lays out the foundations of neuromorphic deep learning machines, and paves the way for learning with streaming spike-event data in neuromorphic platforms at artificial neural network proficiencies.
We demonstrate this in simulation of a custom digital neuromorphic processor using fixed point, discrete-time dynamics.

\section{Results}
\subsection{Event-driven Random Backpropagation}\label{sec:supervised}
  \begin{figure*}
    \begin{center}
      \includegraphics[width=\largefigwidth] {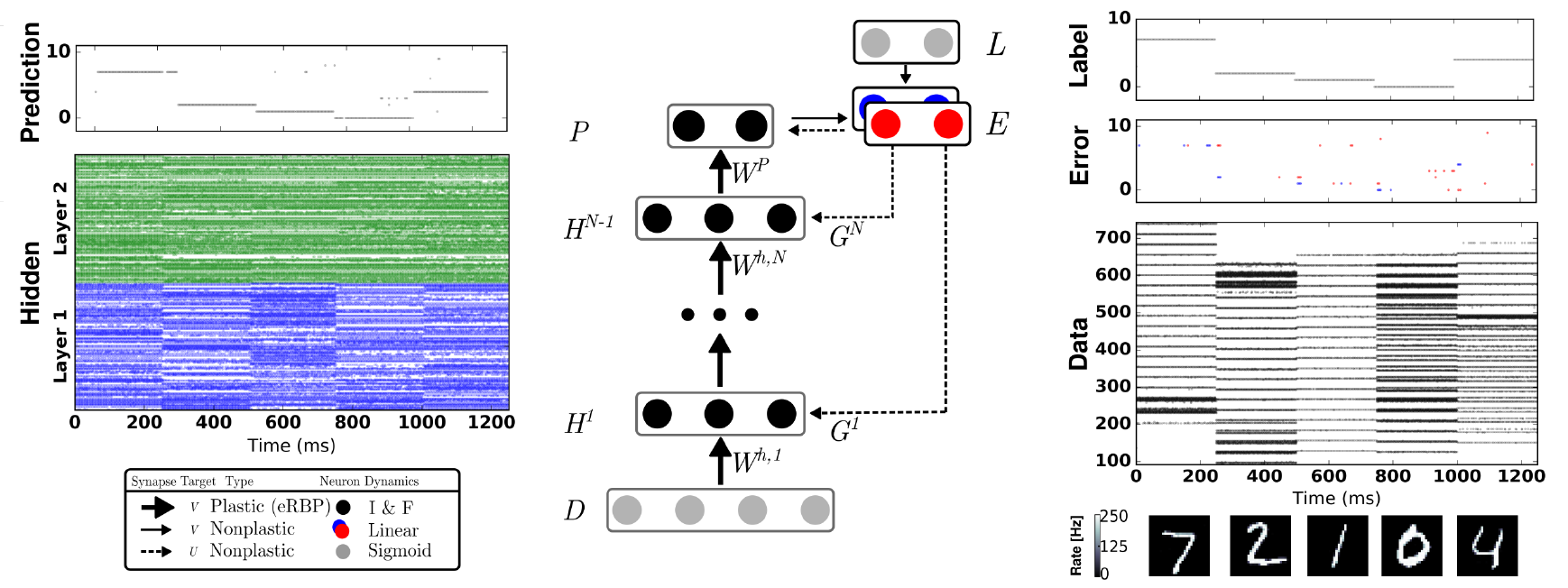}
      \hspace{3pt}
    \end{center}
    \caption{
      \label{fig:erbp} 
      \emph{Network Architecture for Event-driven Random Backpropagation (eRBP) and example spiking activity after training a 784-200-200-10 network for 60 epochs}.
    The network consists in feed-forward layers ($H^1$, ..., $H^{N}$) for prediction and feedback layers for supervised training with labels (targets) $L$.
    Full arrows indicate synaptic connections, thick full arrows indicate plastic synapses, and dashed arrows indicate synaptic plasticity modulation.
    In this example, digits 7,2,1,0,4 were presented in sequence to the network. 
    The digit pixel values are transformed to spike trains (layer D) using a Spike Response Model (\refeq{eq:refr_exp_hazard}). 
    Neurons in the network indicated by black circles were implemented as two-compartment leaky \ac{IF} neurons (\refeq{eq:hidden-neurons} and \refeq{eq:prediction-neurons}).
    The error is the difference between labels (L) and predictions (P), as is implemented using a pair of neurons coding for positive error (blue) and negative error (red), following \refeq{eq:error-coding-neurons}.
    Each hidden neuron receives inputs from a random combination of the pair of error neurons to implement random \ac{BP}.
    Output neurons receives inputs from the pair of error neurons in a one-to-one fashion.
    At the moment of data sample (digit) transitions, bursts of activity (about 3 spikes) in the error neurons occur.
    To prevent the perturbation of the weights during these transitions, no weight updates were undertaken immediately after changing data sample.}
\end{figure*}

%
%

    The central contribution of this article is event-driven RBP (eRBP), a presynaptic spike-driven plasticity rule modulated by top-down errors and gated by the state of the postsynaptic neuron.
    The idea behind this additional modulation factor is motivated by supervised gradient descent learning in artificial neural networks and biologically plausible models of three-factor plasticity rules \cite{Urbanczik_Senn14}, which was argued to subserve supervised, unsupervised and reinforcement learning, an idea that was also reported in \cite{Lillicrap_etal16}. 
    For a given neuron, the eRBP learning dynamics for synapse $k$ can be summarized as follows: 
    \begin{equation}\label{eq:edp}
        \Delta w_{k} = T[t]\,\Theta(I[t]) \mathrm{S^{pre}_k}[t]
    \end{equation}
    where $S_k^{pre}[t]$ represents the spike train of presynaptic neuron $k$ ($S^{pre}_k[t] =1$ if pre-synaptic neuron $k$ spiked at time step $t$), and $\Theta$ is the derivative of the spiking neuron's activation function evaluated at the total synaptic input $I[t]$.     
    The essence of eRBP is contained in the factor $T$. 
    For the final output (prediction) layer, $T$ is equal to the classification error $e$ of the considered neuron, similarly to the delta rule.
    For hidden layers, the $T$ is equal to the error projected randomly to the hidden neurons, \emph{i.e.} $T = \sum_j g_{j} e_j$. 
    where the $g_j$ are random weights that are fixed during the learning.\\
    We found that a boxcar function in place of $\Theta$ provides very good results, while being more amenable to hardware implementation compared to the alternative of computing the exact derivative of the activation function. 
    This choice is motivated by the fact that the activation function of leaky I\&F neurons with absolute refractory period can be approximated by a linear threshold unit with saturation whose derivative is exactly the boxcar function.
    Using a boxcar function with boundaries $b_{min}$ and $b_{max}$, the eRBP synaptic weight update consists of additions and comparisons only, and can be captured using the following operations for each neuron:
    \begin{algorithmic}
      \Function{eRBP}{} 
        \For{$k \in \{$presynaptic spike indices $\mathbf{S}^{pre}$$\}$}
    \If {$ b_{min} < I < b_{max} $}
        $w_k \leftarrow w_k + T$,
    \EndIf
    \EndFor
    \EndFunction
    \end{algorithmic}
    where $\mathbf{S}_{pre}$ is the list of presynaptic neuron indices that have spiked, $T$ is the linear combination of the error vector.
    In the spiking network, $T$ is equal to the voltage of a auxiliary compartment that integrates spikes from the error neurons.
    Provided the second compartment dynamics, no multiplications are necessary for an eRBP update.
    Furthermore, the second compartment dynamics can be made multiplication free (see Methods) given custom bit shift operations and can be disabled after learning.
    Although eRBP is presented using a discrete-time notation, it is straightforward to generalize it in a continuous-time framework.
    This rule is reminiscent of membrane voltage-based rules, where spike-driven plasticity is induced only when membrane voltage is inside an eligibility window \cite{Brader_etal07,Chicca_etal13}.

    The realization of eRBP on neuromorphic hardware requires an auxiliary learning variable for integrating and storing top-down error signals during learning, which can be substantiated as a dendritic compartment.
    Provided this variable, each synaptic weight update incurs only two comparison operations and one addition. 
    Additions and comparisons can be implemented very naturally in neuromorphic VLSI circuits \cite{Liu_etal02}, and costs in the order of tens of femtojoules in digital circuits ($\unit[45]{nm}$ processes \cite{Horowitz14}).
    In a large enough network, the cost of the second compartment dynamics is small compared to the cost of the synaptic update, since for $N$ neurons, there are $N^2$ synapses.
    As a concrete example we use leaky, two compartment, current-based Integrate-and-Fire neurons with additive and multiplicative noise and linear synapses (See Methods).
    The gating term $\Theta$, implemented as two comparisons, operates on the total synaptic input, rather than membrane or calcium as used in other work. 
    This choice is guided by the gradient descent rule, which dictates that the derivative should be evaluated on the total input (Methods). 
    The linearity of the synaptic dynamics allows to use a single dynamical variable for all synapses, such that the value of this dynamical variable is exactly equal to the total synaptic input $I$, and thus readily available at the neuron and the synapses.
    
    The eRBP rule combined with its ability to learn deep representations with near equal accuracies as described below can enable neuromorphic deep learning machines on a wide variety of tasks.
    In the following, we focus on a design that is tailored for digital neuromorphic design, namely that some non-plastic synaptic weights can be exactly matched. 
    Its implementation in a mixed signal design prone to fabrication mismatch and other non-idealities is the subject of ongoing work.

    \subsection{Spiking Networks Equipped with eRBP Learn with High Accuracy}
    We demonstrate eRBP in networks consisting of one and two hidden layers trained on permutation invariant MNIST (\reftab{fig:erbp}) , although eRBP can in theory generalize to other datasets, tasks and network architectures as well.
    Rather than optimizing for absolute classification performance, we compare to equivalent artificial (non-spiking) neural networks trained with RBP and standard BP, with free parameters fine-tuned to achieve the highest accuracy on the considered classification tasks (\reftab{tab:erbp}).
    On most network configurations eRBP achieved performances equivalent to those achieved with RBP in artificial networks. 
    When equipped with probabilistic connections (peRBP) that randomly blank-out presynaptic spikes, the network performed better overall. 
    This is because, as learning progresses, a significant portion of the neurons tend to fire near their maximum rate and synchronize their spiking activity across layers as a result of large synaptic weights (and thus presynaptic inputs).
    Synchronized spike activity is not well captured by our rate model, which is assumed by the eRBP (see Methods).
    Additive noise has relatively small effect when the magnitude of the presynaptic input is large.
    However, multiplicative blank-out noise improves learning by introducing irregularity in the presynaptic spike-trains even when presynaptic neurons fire regularly. 
    Our previous work \cite{Neftci_etal16} suggested that the probabilistic connections implement DropConnect regularization \cite{Wan_etal13}.
    In contrast with \cite{Wan_etal13}, the probabilistic connections remain enabled both during learning and inference because the network dynamics depend strongly on this stochasticity.
    Interestingly, this type of ``always-on'' stochasticity also was argued to approximate Bayesian inference with Gaussian processes \cite{Gal_Ghahramani15}.

    Overall, the learned classification accuracy is comparable with those obtained with offline training of spiking neural networks (\emph{e.g.} GPUs) using standard BP.

    Transitions between two data samples of different class (digit) are marked by bursts of activity in the error neurons (\reffig{fig:erbp}).
    To overcome this problem, weight updates were disabled the first $\unit[50]{ms}$ after the new digit onset.
    In future work involving practical applications on autonomous systems, it will be beneficial to interleave learning and inference stages without explicitly controlling the learning rate. 
    One way to achieve this is to introduce a negative bias in the error neurons by means of a constant negative input and an equal positive bias in the label neurons such that the error neuron can be only be active when an input label is provided\footnote{Such logical ``and'' operation on top of a graded signal was previously used for conditional signal propagation in neuromorphic VLSI spiking neural networks \cite{Neftci_etal13}.}.
    The same solution can overcome the perturbations caused by bursts of error activity during digit transitions (see red and blue spikes in \reffig{fig:erbp}).

    The presence of these bursts of error activity suggest that eRBP could learn spatiotemporal sequences as well. 
    However, learning useful latent representations of the sequences requires solving a temporal credit assignment problem at the hidden layer -- a problem that is commonly solved with gradient \ac{BP}-through-time in artificial neural networks \cite{Rumelhart_etal88} -- which could be tackled using synaptic eligibility dynamics based on ideas of reinforcement learning \cite{Sutton_Barto98}.
    \begin{figure}
          \centering \includegraphics[width=\largefigwidth] {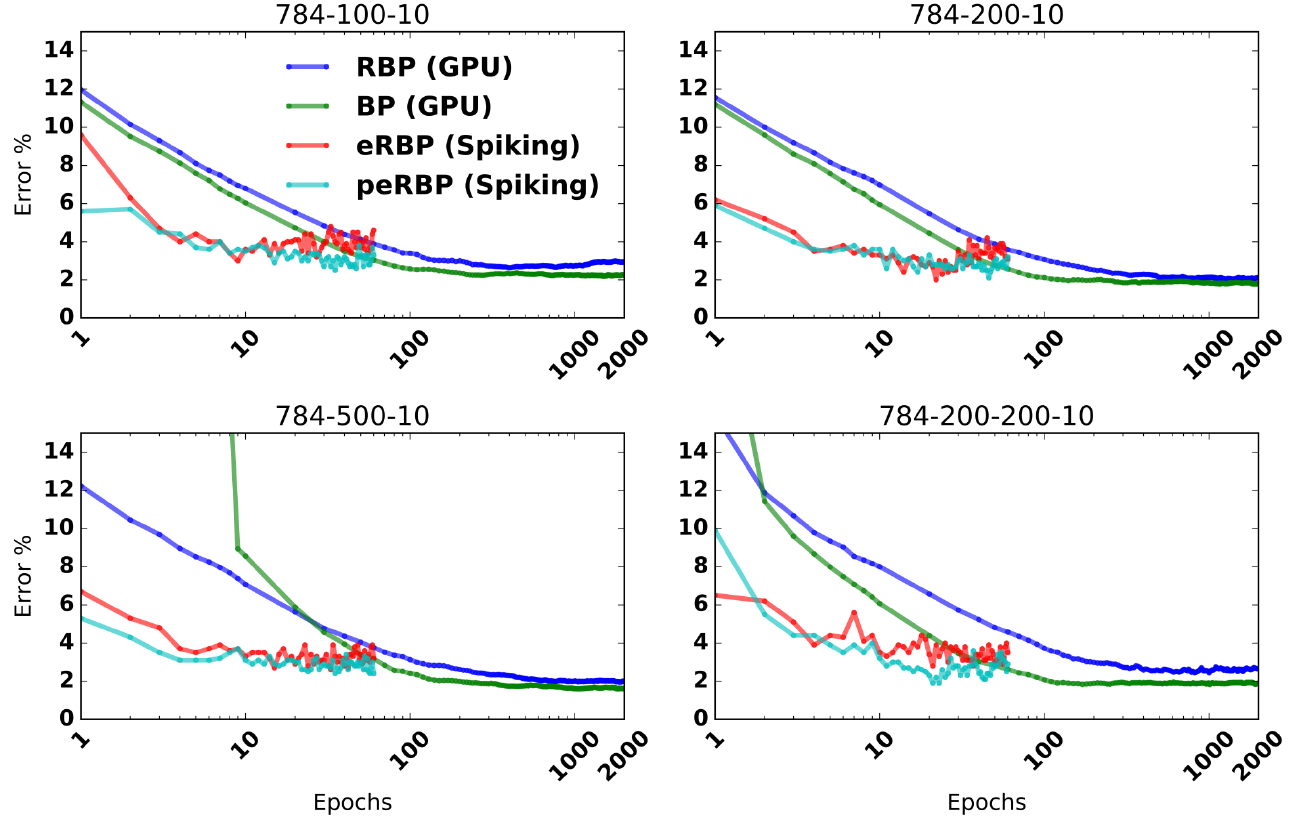}
          \caption{\label{fig:edp} MNIST Classification error on fully connected artificial neural networks (BP and RBP) and on spiking neural networks (eRBP and peRBP).}
    \end{figure}
\begin{table*}

  \centering
  \begin{tabular}{l l l l l}  
    \toprule
    \textbf{Network}&\multicolumn{4}{c}{\textbf{Classification Error}}\\
    & 
    eRBP & 
    peRBP & 
    RBP (GPU) & 
    BP (GPU)\\ 
    \midrule                  
    784-100-10     &\textbf{3.94\%} &\textbf{3.02\%} &2.74\% &2.19\% \\  
    784-200-10     &\textbf{3.13\%} &\textbf{2.51\%} &2.15\% &1.81\% \\  
    784-500-10     &\textbf{2.96\%} &\textbf{2.35\%} &2.08\% &1.8\%\\ 
    784-200-200-10 &\textbf{3.04\%} &\textbf{2.25\%} &2.42\% &1.91\%\\ 
    \bottomrule
  \end{tabular}\\ 
  \caption{\label{tab:erbp} Classification error on the permutation invariant MNIST task (test set). Bold indicates online trained spiking network}
\end{table*}

    \subsection{Classification with Single Spikes is Highly Accurate and Efficient}
\begin{figure}
          \centering \includegraphics[width=\figwidth] {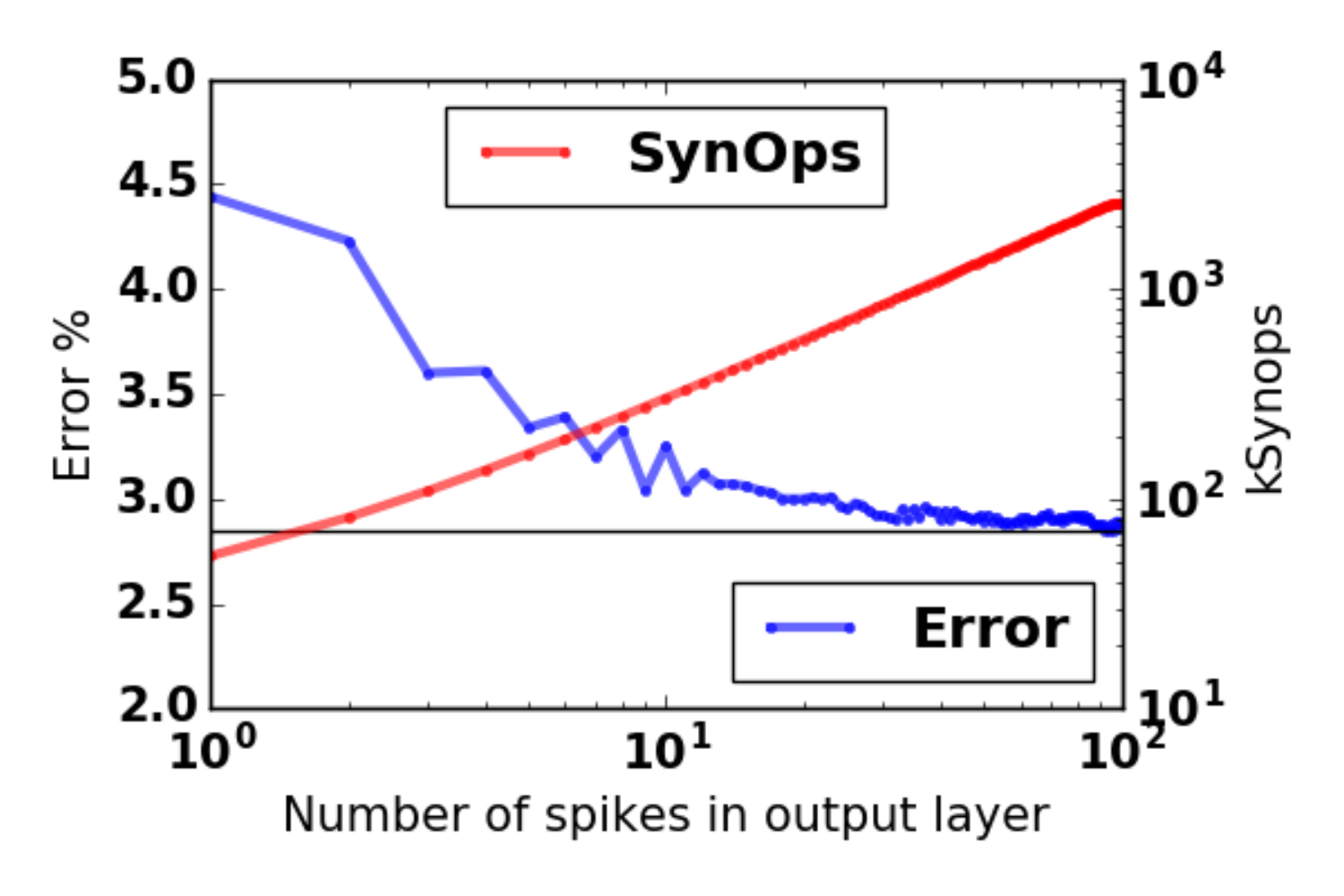}
          \caption{\label{fig:edp-spike} Classification error in the 784-200-10 network as a function of the number of spikes in the prediction layer, and total number of synaptic operations incurred up to each output spike. Horizontal line is 2.85\%. To obtain this data, the network was first stimulated with random patterns, and the spikes in the output layer were counted after $\tau_{syn}=\unit[4]{ms}$.}
    \end{figure}

    The response of the 784-200-10 network after stimulus onset is about one synaptic time constant. 
    Using the first spike after $2\tau_s = 8\mathrm{ms}$ from the stimulus onset for classification leads to about 5\% error (\reffig{fig:edp-spike}), and improves steadily as the number output layer spikes increase.

    In this example, classification using the first spike incurred about $\unit[100]{k}$ synaptic operations (averaged over 10000 test samples), most of which occur between the data and the hidden layer (784 neurons and 200 neurons respectively).
    In existing dedicated neuromorphic hardware \cite{Park_etal14a,Merolla_etal14,Qiao_etal15}, the energetic cost of a synaptic operation is about 20pJ.
    On such hardware, single spike classification in eRBP trained networks can potentially result in about 2$\mu J$ energy per classification. This figure is comparable to the state-of-the-art in digital neuromorphic hardware ($\cong\unit[2]{\mu J}$ at this accuracy \cite{Esser_Etal15}) and current GPU technology ($>\unit{mJ}$).
    We note that no sparsity criterion was enforced in this network. 
    We expect that sparsity implemented explicitly using weight regularization or implicitly using Dropout or DropConnect techniques \cite{Baldi_Sadowski13} can further reduce this energy, by virtue of the lower activity in the hidden layer.

    The low latency response with high accuracy may seem at odds with the inherent firing rate code underlying the network computations (See Methods). 
    However, a code based on the time of the first-spike is consistent with a firing rate code, since a neuron with a high firing rate is expected to fire first \cite{Gerstner_Kistler02}.
    In addition, the onset of the stimulus provokes a burst of synchronized activity, which further favors the rapid onset of the prediction response.
    These results suggest that despite the underlying firing rate code, eRBP can take advantage of the spiking dynamics, with classification accuracies comparable to spiking networks trained exclusively for single-spike classification \cite{Mostafa16}.
    \subsection{Spiking Networks Equipped with eRBP Learn Rapidly and Efficiently}
    The spiking neural network requires fewer iterations of the dataset to reach the peak classification performance compared to the artificial neural network trained with batch gradient descent (\reffig{fig:edp}).
In minibatch learning, weight updates are averaged across the minibatch.
    Batch or Minibatch learning improves learning speed in conventional hardware thanks to vectorization libraries or efficient parallelization with GPUs' SIMD architecture, and lead to smoother convergence.
    However, this approach result in $n_{batch}$ times fewer weight updates per epoch compared to online gradient descent.
    In contrast, the spiking neuron network is updated after each sample presentation, accounting in large part for the faster convergence of learning.
    Other spiking networks trained online using stochastic gradient descent achieved comparable speedup \cite{OConnor_Welling16,Lee_etal16}.
    These results are not entirely surprising since seminal work in stochastic gradient descent established that, with suitable conditions on the learning rate, the solution to a learning problem obtained with stochastic gradient descent is asymptotically as good as the solution obtained with batch gradient descent \cite{Le-Cun_Bottou04} for a given number of samples.
    Furthermore, for equal computational resources, online gradient descent can process more data samples \cite{Le-Cun_Bottou04}, while requiring less memory for implementation.
    Thus, for an equal number of compute operations per unit time, online gradient descent converges faster than batch learning.
    Standard artificial neural networks can be trained using $n_{batch}=1$, but learning becomes much slower on standard platforms because the operations cannot be vectorized across data samples.
    (The converse is not true, however: $n_{batch}>1$ in spiking networks is non-local because it requires storing synaptic weight gradients.)
    It is fortunate that synaptic plasticity is an inherently ``online'' in the machine learning sense, given that potential applications of neuromorphic hardware often involve real-time streaming data. 
    \subsection{eRBP can Learn with low Precision, Fixed Point Representations}
\begin{figure}
      \includegraphics[width=\largefigwidth] {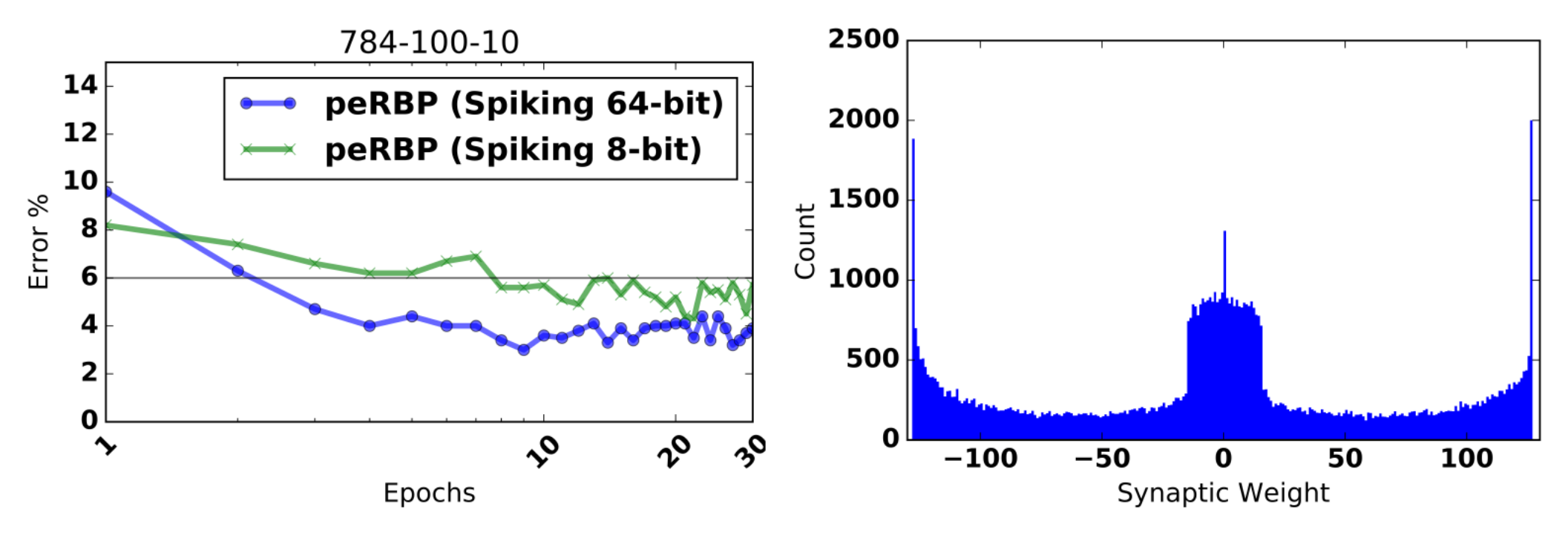}
      \caption{\label{fig:qedp}(\textbf{Left}) MNIST Classification error using a fully connected 784-100-10 network with quantized synaptic weights and neural states (8 bits and 16 bits respectively) (\textbf{Right}). Histogram of synaptic weights of the quantized network after training.}
    \end{figure}

    The effectiveness of stochastic gradient descent degrades when the precision of the synaptic weights using a fixed point representation is smaller than 16 bits \cite{Courbariaux_etal14}.
    This is because quantization determines the smallest learning rate and bounds the range of the synaptic weights, thereby preventing averaging the variability across dataset iterations.
    The tight integration of memory with computing circuits as pursued in neuromorphic chip design is challenging due to space constraints and memory leakage.
    For this reason, full precision (or even 16 bit) computer simulations of spiking networks may be unrepresentative of performance that can be attained in dedicated neuromorphic designed due to quantization of neural states and parameters, and synaptic weights.\\
    Extended simulations suggest that the random BP performances at 10 bits precision is indistinguishable from unquantized weights  \cite{Baldi_etal16}, but whether this is the case for online learning was not yet tested.
    Here, we hypothesize that 8 bit synaptic weight is a good trade-off between the ability to learn with high accuracy and the cost of implementation in hardware.
    To demonstrate robustness to such constraints, we simulate quantized versions of the eRBP network using low precision fixed point representations (8 bits per synaptic weight and 16 bits for neural states).
    Consistent with existing findings, our simulations of eRBP in a quantized 784-100-10 network show that eRBP still performs reasonably well under these conditions (\reffig{fig:qedp}).
    While many weights aggregate at the boundaries, a majority of them remain away from the boundaries.    
    Although the learned accuracies using quantized simulations fall slightly short of the full precision ones, we emphasize that no specific rounding mechanisms \cite{Muller_Indiveri16} was used to obtain these results and are expected to tighten this gap.
\section{Discussion}
The gradient descent \ac{BP} rule is a powerful algorithm that is ubiquitous in deep learning, but when implemented in a von Neumann or neural architecture, it relies on the immediate availability of network-wide information stored with high-precision memory.
More specifically, \cite{Baldi_etal16} and \cite{Lee_etal16} list several reasons why the following requirements of gradient \ac{BP} make them biologically implausible.
The essence of these difficulties is that gradient \ac{BP} is non-local in space and in time when implemented on a neural substrate, and requires precise linear and non-linear computations. 
The feedback alignment work demonstrated that symmetric weights were not necessary for communicating error signals across layers. 
Here we demonstrated a learning rule inspired by Building on feedback alignment \cite{Lillicrap_etal16}, and membrane voltage-gated plasticity rules, and three-factor synaptic plasticity rules proposed in the computational neuroscience literature.
With an adequate network architecture, we find that the spike-based computations and the lack of general linear and non-linear computations and alternating forward and backward steps does not prevent accurate learning. 
Although previous work overcome some of the non-locality problems of gradient \ac{BP} \cite{OConnor_Welling16,Lee_etal16,Lillicrap_etal16}, eRBP overcomes all of the key difficulties using a simple rule that incurs one addition and two comparisons per synaptic weight update.

Taken together, our results suggest that general-purpose deep learning using streaming spike-event data in neuromorphic platforms at artificial neural network proficiencies is realizable.
To emphasize this, we have implemented eRBP using a software simulations using fixed point, discrete-time dynamics, called the neural and synaptic array transceiver.
This simulator is compatible with a digital neuromorphic hardware currently in development, the full details of which will be published elsewhere.

Our experiments target digital implementations of spiking neural networks with embedded plasticity. 
However, membrane-voltage based learning rules implemented in mixed-signal neuromorphic hardware \cite{Qiao_etal15,Huayaney_etal16} are compatible with eRBP provided that synaptic weight updates can be modulated by an external signal on a neuron-to-neuron basis.
Following this route, and combined with the recent advances in neuromorphic engineering and emerging nanotechnologies, eRBP can become key to ultra low-power processing in space and power constrained platforms. 

\subsection{Why Neuromorphic Learning Machines?}
Spiking neural networks, especially those based on the \ac{IF} neuron types severely restrict the possible computations during learning and inference.
With the wide availability of graphical processing units and future dedicated machine learning accelerators, the neuromorphic spike-based approach more machine learning tasks is often heavily criticized as being misguided.
While this is true for some cases and on metrics based on absolute accuracy at some standardized benchmark task, the premise of neuromorphic engineering, \emph{i.e.} that electronic and biological share similar constraints on communication, power and reliability \cite{Mead90}, extend to the algorithmic domain.
That is, accommodating machine learning algorithms within the constraints ultra-low power hardware for adaptive behavior (\emph{i.e.} embedded learning) is likely to result in solutions for communication, computations and reliability that are in strong resemblance with how the brain solves similar problems.
In addition, the neuromorphic approach offers a few advantages over straight artificial neural networks: 1) Asynchronous, event-based communication most often used in neuromorphic hardware considerably reduce the communication between distributed processes, 2) Spiking networks naturally exploit ``rate'' codes and ``spike'' codes where single spikes are meaningful, leading to fast and thus power-efficient and gradual responses (\reffig{fig:edp-spike}, see also \cite{OConnor_Welling16}).  

One prominent example is the Binarized Neural Network (BNN) \cite{Courbariaux_Bengio16}.
The BNN is trained such that weights and activities are -1 or 1, which considerably reduces the energetic footprint of inference, because multiplications are not necessary and the memory requirements for inference are much smaller.
The discrete, quantized dynamics used in this work and developed independently from the BNN shares many similarities, such as binary activations (spikes), low-precision variables, and straight-through gradient estimators.
Our neurally inspired approach innovates several new features for binarized networks: network activity is sparse and data-driven (asynchronous), random variables for stochasticity are generated only when neurons spike, errors are backpropagated only for misclassified examples, and learning is ongoing leading to accurate early single spike classification. 
Many other examples that led to the unprecedented success in machine learning were discovered independently of equivalent neural mechanisms, such as normalization techniques for improving deep learning \cite{Ioffe_Szegedy15,Ren_etal16}, attention, short-term memory for learning complex tasks \cite{Graves_etal14}, and memory consolidation through fast replays for reinforcement learning \cite{Mnih_etal15,Kumaran_etal16}. 
The convergence between the two approaches (neuromorphic \emph{vs.} artificial) will not only improve the design of neuromorphic learning machines, but can also widen the breadth of knowledge transfer between computational neuroscience and deep learning.

\subsection{Relation to Prior Work in Random Backpropagation}
Our learning rule builds on the feedback alignment learning rule proposed in \cite{Lillicrap_etal16}, showing that random feedback can deliver useful teaching signals by aligning the feed-forward weights with the feed-back weights.
The authors also demonstrated a spiking neural network implementing feedback alignment, demonstrating that feedback alignment is able to
implicitly adapt to random feedback when the forward and backward pathways both operate continuously. 
However, their learning rule is not event-based as in eRBP, but operates on a continuous-time fashion that is not directly compatible with spike-driven plasticity, and a direct neuromorphic implementation thereof would be inadequate due to the high bandwidth communication required between neurons.
Furthermore, their model is a spike response model that does not emulate the physical dynamics of spiking neurons such as \ac{IF} neurons.
Another difference between eRBP and the network presented in \cite{Lillicrap_etal16} is that eRBP contains only one error-coding layer, whereas feedback alignment contains one error-coding layer per hidden layer. 
Such direct feedback alignment was recently proposed in \cite{N-kland16} and \cite{Baldi_etal16}, and a theoretical analysis there showed that gradient computed in this fashion point is within 90 degrees of the backpropagated gradient.
\cite{Baldi_etal16} studied feedback alignment in the framework of local learning and the learning channel, and derived several other flavors of random \ac{BP} such as adaptive, sparse, skipped and indirect RBP, along with their combinations.
In related work, \cite{Lee_etal14} showed how feedback weights can be learned to improve the classification accuracy by training the feedback weights to learn the inverse of the feedforward mapping. 

\subsection{Relation to Prior Work in Spiking Deep Neural Networks}
Several approaches successfully realized the mapping of pre-trained artificial neural networks onto spiking neural networks using a firing rate code \cite{OConnor_etal13_real-clas,Cao_etal14_spik-deep,Hunsberger_Eliasmith15_spik-deep,OConnor_Welling16,Esser_etal16,Neftci_etal14_even-cont,Diehl_etal15_fast-high,Das_etal15_gibb-samp,Marti_etal15_ener-neur} 
Such mapping techniques have the advantage that they can leverage the capabilities of existing machine learning frameworks such as Caffe \cite{Jia_etal14_caff-conv} or Theano \cite{Goodfellow_etal13_pyle-mach} for brain-inspired computers.
More recently, \cite{Mostafa16} used a temporal coding scheme where information is encoded in spike times instead of spike rates and the dynamics are cast in a differentiable form.
As a result, the network can be trained using standard gradient descent to achieve very accurate, sparse and power-efficient classification.
Although eRBP achieves comparable results, their approach naturally leads to sparse activity in the hidden layer which can be more advantageous in large and deep networks.

An intermediate approach is to learn online with standard \ac{BP} using spike-based quantization of network states \cite{OConnor_Welling16} and the instantaneous firing rate of the neurons \cite{Lee_etal16}.
\cite{OConnor_Welling16} eschews neural dynamics and instead operates directly on event-based (spiking) quantizations of vectors. 
Using this representation, common neural network operations including online gradient \ac{BP} are mapped on to basic addition, comparison and indexing operations applied to streams of signed spikes.
As in eRBP, their learning rule achieves better results when weight updates are made in an event-based fashion, as this allows the network to update its parameters many times during the processing of a single data sample.
\cite{Lee_etal16} propose a method for training spiking neural networks via a formulation of the instantaneous firing rate of the neuron obtained by low-pass filtering the spikes. 
There, quantities that can be related to the postsynaptic potential (rather than mean rates) are used to compute the derivative of the activity of the neuron, which can provide a useful gradient for backpropagation. 
\cite{Esser_etal16} use multiple spiking convolutional networks trained offline to achieve near state-of-the-art classification in standard benchmark tasks.
Their approach maps onto the all-digital spiking neural network architecture using trinary weights.
For the above approaches, the eRBP learning rule presented here can be used as a drop-in replacement and can reduce the computational footprint of learning by simplifying the backpropagated chain path and by operating directly with locally available variables \emph{i.e.} membrane potentials and spikes.


\subsection{Relation to Prior Work in Spike-Driven Plasticity Rules}
STDP has been shown to be very powerful in a number of different models and tasks \cite{Thorpe_etal01,Nessler_etal13,Neftci_etal14}.
Although the implementation of acausal updates (triggered by presynaptic firing) is typically straightforward in cases where presynaptic lookup tables are used, the implementation of causal updates (triggered by postsynaptic firing) can be challenging due to the requirement of storing a reverse look-up table.
Several approximations of STDP exist to solve this problem \cite{Pedroni_etal16a,Galluppi_etal14}, but require dedicated circuits.

Thus, there is considerable benefit in hardware implementations of synaptic plasticity rules that forego the causal updates.
Such rules, which we referred to as spike-driven plasticity, can be consistent with STDP \cite{Brader_etal07,Sheik_etal16a,Qiao_etal15,Clopath_etal10}, especially when using dynamical variables that are representative of the pre- and postsynaptic firing rates (such as calcium or average membrane voltage).

A common feature among spike-driven learning rules is a modulation or gating with a variable that reflects the average firing rate of the neuron for example through calcium concentration \cite{Graupner_Brunel12,Huayaney_etal16} or the membrane potential \cite{Clopath_etal10,Sheik_etal16a} or both \cite{Brader_etal07}.
\cite{Sheik_etal16a} recently proposed a membrane-gated rule inspired by calcium and voltage-based rules with homeostasis for learning unsupervised spike pattern detection.
Their rule statistically emulates pairwise STDP using presynaptic spike timing only and using additions and multiplications.
Except for homeostasis, eRBP follows similar dynamics but potentiation and depression magnitudes are dynamic and determined by external modulation, and comparisons are made on total synaptic currents to avoid the effect of the voltage reset after firing.

The two compartment neuron model used in this work is motivated from conductance-based dynamics in \cite{Urbanczik_Senn14} and previous neuromorphic realizations of two compartment mixed signal spiking neurons \cite{Park_etal14a}.
Although the spiking network used in this work is current-based rather than conductance-based, eRBP shares strong similarities to the three-factor learning rule employed in \cite{Urbanczik_Senn14}.
The latter is composed of three factors: an approximation of the prediction error, the derivative of the membrane potential with respect to the synaptic weight, and a positive weighting function that stabilizes learning in certain scenarios.
The first factor corresponds to the error modulation, while the second and third factors roughly correspond to the presynaptic activity and the derivative of the activation function.
The differences between eRBP and \cite{Urbanczik_Senn14} (besides from the \emph{random} \ac{BP} which was considered in \cite{Lillicrap_etal16}) stems mainly from two facts: 1) the firing rate description used here for simplicity and for easier comparisons between artificial neural networks and spiking neural networks and 2) eRBP is fully event-based in the sense that weights are updated only when the presynaptic neurons spike, in order to make memory and compute operations more efficient in hardware.

\section{Conclusions and Future Directions}
This article demonstrates a local learning rule for deep, feed-forward neural networks achieving classification accuracies on par with those obtained using equivalent machine learning algorithms. 
The learning rule combines two features: 1) algorithmic simplicity: one addition and two comparisions per synaptic update provided one auxiliary state per neuron and 2) Locality: all the information for the weight update is available at each neuron and the synapse. 
The combination of these two features enables learning dynamics for deep learning in neuromorphic hardware.

Existing literature suggests that that random \ac{BP} also works for unsupervised learning \cite{Lee_etal14,Baldi_etal16} in deeper and convolutional networks. 
It can be reasonably expected that the deep learning community will uncover many variants of random \ac{BP}, including in recurrent neural networks for sequence learning and memory augmented neural networks.
In tandem with these developments, we envision that such RBP techniques will enable the embedded learning of pattern recognition, attention, working memory and action selection mechanisms which promise transformative hardware architectures for embedded computing.

This work has focused on unstructured, feed-forward neural networks and a single benchmark task across multiple implementations for ease of comparison.
Limitations in deep learning algorithms are often invisible on ``small'' datasets like MNIST \cite{Liao_etal15}.
Random \ac{BP} was demonstrated to be effective in a variety of tasks and network structures \cite{Liao_etal15,Baldi_etal16}, including convolutional neural networks.
Although random \ac{BP} was reported to work well in this case \cite{Liao_etal15}, the parameter sharing in convnets is inherently non-local.
Despite this non-locality, neuromorphic implementation of convnets are still possible in neuromorphic \cite{Qiao_etal15} if presynaptic connectivity tables are stored rather than postsynaptic tables.

\section{Methods}

\subsection{Derivation of Event-driven Random Backpropagation}
  In artificial neural networks, the mean-squared cost function for one data sample in a single layer neural network is:
  \begin{equation}
    \begin{split}
      L   &=\frac12  \sum_i e_i^2,\\
      e_i &= (y_i - l_i),
    \end{split}
  \end{equation}
  where $e_i$ is the error of prediction neuron $i$, $y_i = \phi(\sum_j w_{ij} x_j)$ is the activity of the prediction neuron $i$ with activation function $\phi$, $\mathbf{x}$ is the data sample and $l_i$ is the label associated to the data sample.
  The task of learning is to minimize this cost over the entire dataset.
  The gradient descent rule in artificial neural networks is often used to this end by modifying the network parameters $\mathbf{w}$ in the direction opposite to the gradient:
  \begin{equation}\label{eq:bp_shallow}
    \begin{split}
    w_{ij}[t+1] &= w_{ij}[t] - \eta \frac{\mathrm{\partial}}{\mathrm{\partial} w_{ij}} L,\\
    \text{where }\frac{\mathrm{\partial}}{\mathrm{\partial} w_{ij}} L &= \phi'(\sum_j w_{ij} x_j) e_i x_j.
    \end{split}
  \end{equation}
  where $\eta$ is a small learning rate. 
  In deep networks, \emph{i.e.} networks containing one or more hidden layers,  the weights of the hidden layer neurons are modified by backpropagating the errors from the prediction layer using the chain rule:
  \begin{equation}\label{eq:bp_deep}
    \begin{split}
      \frac{\mathrm{\partial}}{\mathrm{\partial} w^{l}_{ij}} L &= \delta_{ij}^{l}  y^{l-1}_j,\\
      \text{where }\delta_{ij}^{l} &= \phi'(\sum_j w^{l-1}_{ij} y^{l-1}_j) \sum_k \delta_{ik}^{l+1}w_{ik}^{l+1}.
    \end{split}
  \end{equation}
  where the $\delta$ for the topmost layer is $e_i$, as in \refeq{eq:bp_shallow} and $y$ at the bottommost layer is the data $x$.
  This update rule is the well-known gradient back propagation algorithm ubiquitously used in deep learning \cite{Rumelhart_etal88}.
  Learning is typically carried out in forward passes (evaluation of the neural network activities) and backward passes (evaluation of the $\delta$s).
  The computation of the $\delta$ requires knowledge of the forward weights, thus gradient \ac{BP} relies on the immediate availability of a symmetric transpose of the network for computing the backpropagated errors $\delta_{ij}^{l}$. 
  Often the access to this information funnels through the von Neumann bottleneck, which dictates the fundamental limits of the computing substrate.    

  In the random \ac{BP} rule considered here, the \ac{BP} term $\delta$ is replaced with: 
  \begin{equation}\label{eq:rbp_deep}
      \delta_{RBP}^{l} = \phi'(\sum_j w^{l-1}_{ij} y^{l-1}_j) \sum_k e_k g_{ik}^l
  \end{equation}
  where $g_{ik}^l$ are fixed random numbers. 
  This backpropagated term does not depend on the previous layer $l+1$, and thus does not have a recursive structure as in standard \ac{BP} (\refeq{eq:bp_deep}) or feedback alignment \cite{Lillicrap_etal16}. 
  This form was previously referred to as direct feedback alignment \cite{N-kland16} or skipped RBP \cite{Baldi_etal16} and was shown to perform equally well on a broad spectrum of tasks. 
  A detailed justification of random \ac{BP} is out of the scope of this article, and interested readers are referred to \cite{N-kland16,Baldi_etal16,Lillicrap_etal16}.

  In the context of models of biological spiking neurons, RBP is appealing because it circumvents the problem of calculating the backpropagated errors and does not require bidirectional synapses or symmetric weights. 
  RBP works remarkably very well in a wide variety of classification and regression problems, using supervised and unsupervised learning in feed-forward networks, with a very small penalty in accuracy.
  
  The above \ac{BP} rules are commonly used in artificial neural networks, where neuron outputs are represented as single scalar variables.
  To derive an equivalent spike-based rule, we start by matching this scalar value is the neuron's instantaneous firing rate.
  The cost function and its derivative for one data sample is then:
  \begin{equation}
    \begin{split}
      L_{sp} &= \frac12 \sum_i (\nu^p_i(t) - \nu^l_i(t))^2\\
      \frac{\mathrm{\partial}}{\mathrm{\partial} w_{ij}} L_{sp} &= \sum_i e_i(t) \frac{\mathrm{\partial}}{\mathrm{\partial} w_{ij}} \nu^p_i(t)
    \end{split}
  \end{equation}
  where $e_i(t)$ is the error of prediction unit $i$ and $\nu^p$, $\nu^l$ are the firing rates of prediction and label neurons, respectively.

  Random \ac{BP} (\refeq{eq:rbp_deep}) is straightforward to implement in artificial neural network simulations. 
  However, spiking neurons and synapses, especially with the dynamics that can be afforded in low-power neuromorphic implementations typically do not have arbitrary mathematical operations at their disposal. 
  For example, evaluating the derivative $\phi$ can be difficult depending on the form of $\phi$ and multiplications between the multiple factors involved in RBP can become very costly given that they must be performed at every synapse for every presynaptic event.

  In the following, we derive an event-driven version of RBP that uses only two comparisons and one addition for each presynaptic spike to perform the weight update.
  The derivation proceeds as follows: 1) Derive the firing rate $\nu$, \emph{i.e} the equivalent of $\phi$ in the spiking neural network, 2) Compute its derivative $\frac{\mathrm{\partial}}{\mathrm{\partial} w_{ij}} \nu_i(t)$, 3) Introduce modulation with a random linear combination of the classification error to the hidden neurons, 4) Devise a plasticity rule that increments the weight with the product of the latter two factors times the presynaptic activity.

  \subsubsection*{Activation Function of Spiking Neurons with Background Poisson Noise and its Derivative}\label{sec:methods_ssm_math}
     The dynamics of spiking neural circuits driven by Poisson spike trains is often studied in the diffusion approximation \cite{Wang99_syna-basi,Brunel_Hakim99_fast-glob,Brunel00_dyna-spar,Fusi_Mattia99_coll-beha,Renart_etal03_comp-neur,Deco_etal08_dyna-brai,Tuckwell05_intr-to}.
     In this approximation, the firing rates of individual neurons are replaced by a common time-dependent population activity variable with the same mean and two-point correlation function as the original variables, corresponding here to a Gaussian process.
   The approximation is true when the following assumptions are verified:
     1) the charge delivered by each spike to the postsynaptic neuron is small compared to the charge necessary to generate an action potential,
     2) the number of afferent inputs to each neuron is large,
     3) the spike times are uncorrelated.
   In the diffusion approximation, only the first two moments of the synaptic current are retained.
   The currents to the neuron, $I(t)$, can then be decomposed as:
   \begin{equation}\label{eq:diff_approx_eq}
       I(t) = \mu + \sigma \eta(t),
   \end{equation}
   where $\mu = \langle I(t) \rangle = \sum_j w_j \nu_j $ and $\sigma^2 = w_{bg}^2 \nu_{bg}$, where $\nu_{bg}$ is the firing rate of the background activity, and  $\eta(t)$ is the white noise process. 
   We restrict neuron dynamics to the case of synaptic time constants that are much larger than the membrane time constant, \emph{i.e.} $\tau_m\ll \tau_{syn}$, such that we can neglect the fluctuations caused by synaptic activity from other neurons in the network \emph{i.e.} $\sigma$ is constant.
   Although the above dynamics are not true in general, in a neuromorphic approach, the parameters can be chosen accordingly during configuration or at design.

   In this case, the neuron's membrane potential dynamics is an Ornstein-Uhlenbeck (OU) process \cite{Gardiner12_hand-stoc}.
   The stationary distribution of the freely evolving membrane potential (no firing threshold) is a Gaussian distribution:
   \begin{equation}\label{eq:v_distr}
     V_{nt} \sim N(\frac{\mu}{g_L},\frac{\sigma^2}{2 g_L^2 \tau_{m}}).
   \end{equation}
   where $g_L$ is the leak conductance and $\tau_m$ is the membrane time constant.
   Although this distribution is generally not representative of the membrane potential of the \ac{IF} neuron due to the firing threshold \cite{Gerstner_Kistler02}, the considered case $\tau_m\ll \tau_{syn}$ yields approximately a truncated Gaussian distribution, where neurons with $V_{nt}>0$ fire at their maximum rate of $\frac1{\tau_{refr}}$.
 This approximation is less exact for very large $\mu$ due to the resetting, but the resulting form highlights the essence of eRBP while maintaining mathematical tractability. 
 Furthermore, using a first-passage time approach, \cite{Petrovici_etal13} computed corrections that account for small synaptic time constants and the effect of the firing threshold on this distribution.

 The firing rate of neuron $i$ is approximately equal to the inverse of the refractory period, $\nu_i = \tau_{refr}^{-1}$ with probability $P(V_{nt,i} (t+1) \ge 0| \mathbf{s} (t))$ and zero otherwise. 
 The probability is equal to one minus the cumulative distribution function of $V_{nt,i}$:
  \[
    P(V_{nt,i} (t+1) \ge 0| \mathbf{s} (t)) = \frac12 \left( 1 + \operatorname{erf}\left( \frac{\mu_i(t)}{\sigma_{OU}\sqrt{2}}\right) \right),
  \]
  where ``$\mathrm{erf}$'' stands for the error function.
  The firing rate of neuron $i$ becomes:
   \begin{equation}                      
       \begin{split}
         \nu_i & = \frac1{\tau_{refr}}\frac12\left(1+\operatorname{erf}\left( \frac{\sqrt{\tau_{m}}}{\sigma}{\sum_j w_{ij} \nu_j}  \right)\right).
       \end{split}
   \end{equation}

   For gradient descent, we require the derivative of the neuron's activation function with respect to the weight $w$. 
   By definition of the cumulative distribution, this is the Gaussian function in \refeq{eq:v_distr} times the presynaptic activity:
   \begin{equation}                      
       \begin{split}
         \frac{\partial}{\partial w_{ij}} \nu_i & \propto \frac{1}{\sigma_{OU} \sqrt{2\pi}} \exp(-\frac{\mu_i(t)^2}{2\sigma_{OU}^2}) \nu_j(t).
       \end{split}
   \end{equation}
   As in previous work \cite{Neftci_etal14}, we replace $\nu_j(t)$ in the above equations with the presynaptic spike train $s_{j}(t)$ to obtain an asynchronous, \emph{event-driven} update, where the derivative is evaluated only when the presynaptic neuron spikes.
   This approach is justified by the fact that the learning rate is typically small, such that the event-driven updates are averaged at the synaptic weight variable \cite{Gerstner_Kistler02}.
   Thus the derivative becomes:
   \begin{equation}\label{eq:deriv}
       \begin{split}
         \frac{\partial}{\partial w_{ij}} \nu_i & \propto 
                    \begin{cases}
                      \exp(-\frac{\mu_i(t)^2}{2\sigma_{OU}^2}) \text{ if $s_j(t)=1$} \\
                      0 \text{ otherwise}
                    \end{cases}.
       \end{split}
   \end{equation}

   In the considered spiking neuron dynamics, the Gaussian function is not directly available. 
   Although, a sampling scheme based on the membrane potential to approximate the derivative is possible, here we follow a simpler solution:
   Backed by extensive simulations, and inspired by previously proposed learning rules based on membrane potential gated learning rules \cite{Sheik_etal16a,Brader_etal07,Clopath_etal10}, we find that replacing the Gaussian function with a boxcar function $\Theta$ operating on the total synaptic input, $I(t)$, with boundaries $b_{min}$ and $b_{max}$ yields results that are as good as using the exact derivative.    
   With appropriate boundaries, $\Theta(I(t))$ can be interpreted as a piecewise constant approximation of the Gaussian function\footnote{or equivalently, for the purpose of the derivative evaluation, the activation function is approximated as a rectified linear with hard saturation at $\tau_{refr}^{-1}$, also called ``hard tanh'' in the machine learning community.} since $I(t)$ is proportional to its argument $\sum_j w_{ij} \nu_j$, and has the advantage that an explicit multiplication with the modulation is unnecessary in the random \ac{BP} rule (explained below).    
   \begin{equation}                      
       \begin{split}
         \frac{\partial}{\partial w_{ij}} \nu_i \propto 
                    \begin{cases}
                      1 \text{ if $s_j(t)=1$ and $b_{min} < I_i(t) < b_{max}$} \\
                      0 \text{ otherwise}
                    \end{cases}
       \end{split}
   \end{equation}
   The resulting derivative function is similar in spirit to straight-through estimators used in machine learning \cite{Courbariaux_Bengio16}.

  \subsubsection*{Derivation of Event-Driven Random Backpropagation}\label{sec:methods_erbp}

  For simplicity, the error $e_i(t)$ is computed using a pair of spiking neurons with a rectified linear activation function.
  One neuron computes the positive values of $e_i(t)$, while the other neuron computes the negative values of $e_i(t)$ such that:
  \begin{equation}
  \begin{split}
    \nu^{E+}_i(t) &\propto   \nu^p_i(t) - \nu^l_i(t),\\
    \nu^{E-}_i(t) &\propto - \nu^p_i(t) + \nu^l_i(t).
  \end{split}
  \end{equation}
  Each pair of error neurons synapse with a leaky dendritic compartment $U$ of the hidden and prediction neurons using equal synaptic weights with opposite sign, generating a dendritic potential proportional to $(\nu^{E+}_i(t) - \nu^{E-}_i(t)) \cong e_i$.
  Several other schemes for communicating the errors are possible. 
  For example an earlier version of eRBP used on a positively biased error neuron per class (rather than a positive negative pair as above) such that the neuron operated (mostly) in the linear regime.
  This solution led to similar results but was computationally more expensive due to error neurons being strongly active even when the classification was correct.
  Population codes of heterogeneous neurons as in \cite{Salinas_Abbott94,Eliasmith_Anderson04} may provide even more flexible dynamics for learning.
The weight update for the last layer becomes:
   \begin{equation}\label{eq:deriv}
       \begin{split}
           \Delta w_{ij} & \propto 
                    \begin{cases}
                        - e_i(t) \text{ if $s_j^h(t)=1$}\text{ and } b_{min}<I_i<b_{max} \\
                      0 \text{ otherwise}
                    \end{cases}.
       \end{split}
   \end{equation}

   The weight update for the hidden layers is similar, except that a random linear combination of the error is used instead of $e_i$:
   \begin{equation}\label{eq:deriv2}
       \begin{split}
           \Delta w^C_{ij} & \propto 
                    \begin{cases}
                      - \sum_k g_{ik} e_k^E (t) \text{ if $s_j^C(t)=1$}\text{ and } b_{min}<I_i<b_{max} \\
                      0 \text{ otherwise}
                    \end{cases}.
       \end{split}
   \end{equation}
   where $C=\{d,h\}$.
   All weight initializations are scaled with the number of rows and the number of columns as $g_{ik} \sim U(\sqrt{\frac{6}{N_E+N_H}})$ , where $N_E$ is the number of error neurons and $N_H$ is the number of hidden neurons.

    In the following, we detail the spiking neuron dynamics that can efficiently implement eRBP.

    \subsection{Spiking Neural Network and Plasticity Dynamics}
    The network used for eRBP consists of one or two feed-forward layers (\reffig{fig:erbp}) with $N_d$ ``data'' neurons, $N_h$ hidden neurons and $N_p$ prediction neurons.
    The top layer, labeled $P$, is the prediction.
    The feedback from the error population is fed back directly to the hidden layers` neurons.
    The network is composed of three types of neurons: \\
    1) \textbf{Error-coding neurons} are non-leaky \ac{IF} neurons following the linear dynamics:
    \begin{equation}\label{eq:error-coding-neurons}
      \begin{split}
        C \frac{\mathrm{d}}{\mathrm{d}t} V^{E+}_i &= w^{L+} (s_i^P(t) - s_i^L(t))\\
        \text{if } V^{E+}>V_T^{E} &\text{ then } V^{E+}\leftarrow V^{E+}-V_T^{E},
      \end{split}
    \end{equation}
    where $s_i^P(t)$ and $s_i^L(t)$ are spike trains from prediction neurons and labels (teaching signal). 
    To prevent negative runaway dynamics, a rigid boundary at zero is imposed.
    In addition, the membrane potential is lower bounded to  $V_T^{E}$.
    Each error neuron has one counterpart neuron with weights of opposite sign, \emph{i.e.} $w^{L-} = - w^{L+}$ to encode the negative errors.
    The firing rate of the error-coding neurons is proportional to a linear rectification of the inputs. 
    For simplicity, the label spike train is regular with firing rate equal to $\tau_{refr}^{-1}$. 
    When the prediction neurons classify correctly, $(s_i^P(t) - s_i^L(t)) \cong 0$, such that the error neurons remain silent.\\
    2) \textbf{Hidden neurons} follow current-based leaky \ac{IF} dynamics:
    \begin{equation}\label{eq:hidden-neurons}
      \begin{split}
        C \frac{\mathrm{d}}{\mathrm{d}t} \vectwo{V_i^{h}}{U_i^{h}} &= - \vectwo{g_V V_i^{h}}{g_U U_i^{h}} + \vectwo{I_i^h + \sigma_w \eta^h_i(t)}{\sum_{k=1}^{N_{L}} g_{ik}^{E+} s^{E+}_k(t) - g_{ik}^{E-} s^{E-}_k(t)}\\
        \tau_{syn} \frac{\mathrm{d}}{\mathrm{d}t} I_i^h &= -I_i^h + \sum_{k=1}^{N_{d}} w_{ik}^{d} s_k^{d}(t) \xi(t) + \sum_{j=1}^{N_{h}} w_{ij}^{h}  s_j^{h}(t) \xi(t)\\
        \text{if } V_i(t)>V_T & \text{ then } V_i^h\leftarrow 0 \text{ during refractory period }\tau_{refr}.
      \end{split}
    \end{equation}
    where $s_k^d(t)$ and $s_j^h(t)$ are the spike trains of the data neurons and the hidden neurons, respectively, $I^h$ are current-based synapse dynamics, $\sigma_w s^{bg}_i(t)$ a Poisson process of rate $1$kHz and amplitude $\sigma_w$, and $\xi$ is a stochastic Bernouilli process with probability $p$ (indices $i,j$ are omitted for clarity).
    The Poisson process simulates background Poisson activity and contributes additively to the membrane potential, whereas the Bernouilli process contributes multiplicatively by randomly ``blanking-out'' the proportion $(1-p)$ of the input spikes.
    In this work, we consider feed-forward networks, \emph{i.e} the weight matrix $w^{h}$ is restricted to be upper diagonal.
    Each neuron is equipped with a separate ``dendritic'' compartment $U^h_i$ following similar subthreshold dynamics as the membrane potential and where $s^E(t)$ is the spike train of the error-coding neurons and $g^{E}_{ij}$ is a fixed random matrix.
    The dendritic compartment is not directly coupled to the ``somatic'' membrane potential $V^h_i$, but indirectly through the learning dynamics.
    For every hidden neuron $i$, $\sum_j w_{ij}^{E}=0$, ensuring that the spontaneous firing rate of the error-coding neurons does not bias the learning.
    The synaptic weight dynamics follow a dendrite-modulated and gated rule:
    \begin{equation}
          \frac{\mathrm{d}}{\mathrm{d}t} w_{ij}^h = \eta U^{h}_{i} \Theta(I^h_i)s_{j}^{h}(t).
    \end{equation}
    where $\Theta$ is a boxcar function with boundaries $b_{min}$ and $b_{max}$.

    3) \textbf{Prediction neurons}, synapses and synaptic weight updates follow the same dynamics as the hidden neurons except for the dendritic compartment, and one-to-one connection with pairs of error-neurons associated to the same class: 
    \begin{equation}\label{eq:prediction-neurons}
      \begin{split}
        C \frac{\mathrm{d}}{\mathrm{d}t} \vectwo{V_i^{P}}{U_i^{P}} &= - \vectwo{g_V V_i^{P}}{g_U U_i^{P}} + \vectwo{I_i^P + \sigma_w \eta^P_i(t)}{w^E s^{E+}_i(t) - w^E s^{E-}_i(t)}.\\
      \end{split}
    \end{equation}

\begin{figure}
      \begin{center}
          \includegraphics[width=\figwidth] {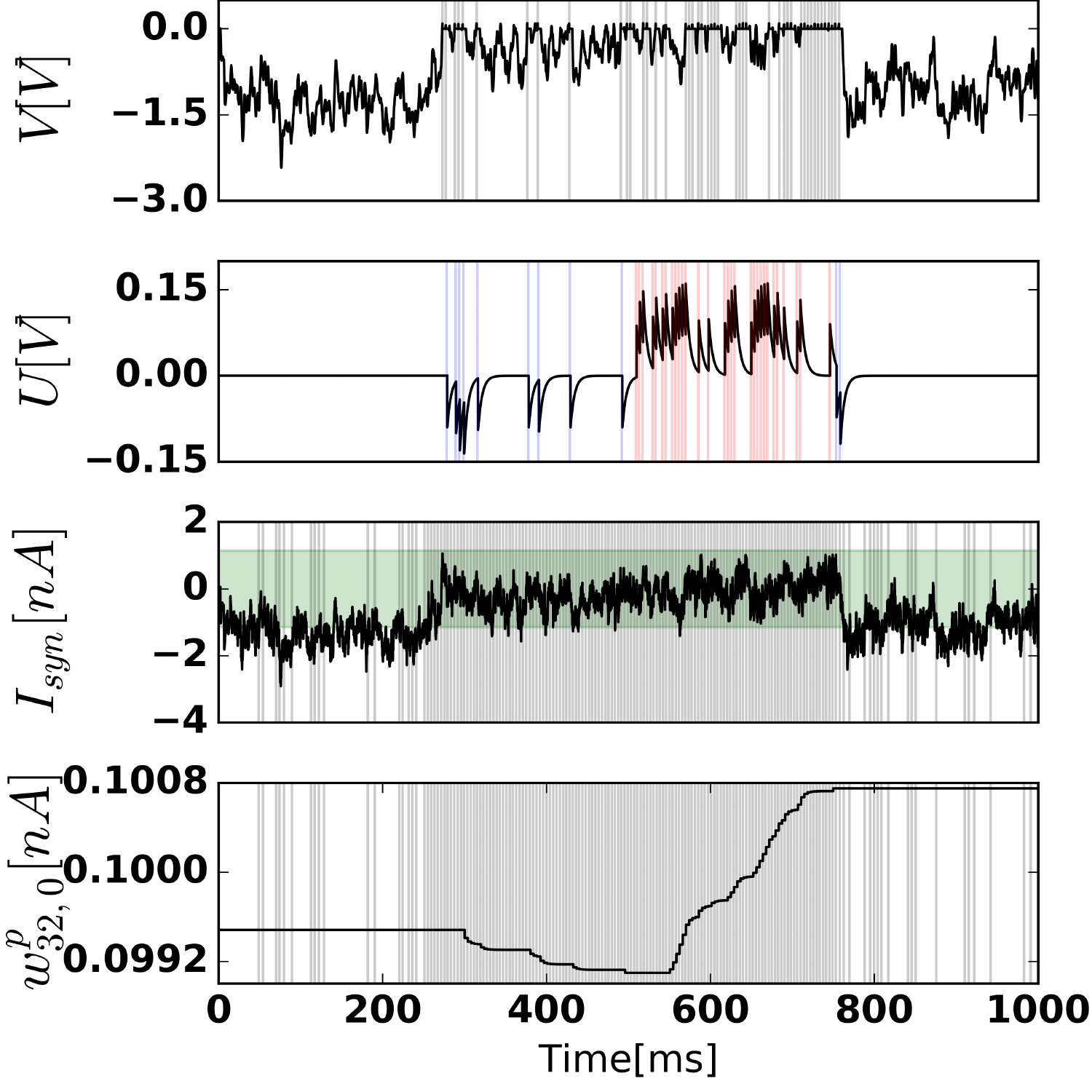}
      \end{center}
          \caption{\label{fig:trace}\emph{Neural states and synaptic weight of the prediction neuron after 500 training examples.} (top) Somatic membrane potential dynamics of prediction neuron $0$, where output spikes are superimposed as grey vertical bars. (Middle-top) Dendritic membrane potential, where blue and red bars indicate negative error neuron 0 spikes and positive error neuron 0 spikes, respectively. In the time range (500,750), the digit 0 is presented to the network. (Middle-bottom) Total synaptic current of prediction neuron $0$, where superimposed vertical bars are presynaptic spikes of hidden neuron $32$. The green shaded area ($b_{min}$, $b_{max}$) corresponds to the plasticity-enabled region, \emph{i.e.} the approximate derivative function $\Theta (I^p(t))$. (Bottom) Synaptic weight between hidden neuron $32$ and  prediction neuron $0$.}
    \end{figure}

    The spike trains at the data layer were generated using a stochastic neuron with instantaneous firing rate (exponential hazard function \cite{Gerstner_Kistler02} with absolute refractory period):
    \begin{equation} \label{eq:refr_exp_hazard}
    \nu^d(d, t-t') =  
      \begin{dcases*}
        0 & if $t-t'<\tau_{refr}$\\
        \tau_{refr}^{-1}\exp(\beta d + \gamma) & $t-t' \geq \tau_{refr}$
      \end{dcases*},
    \end{equation}
    where $d$ is the intensity of the pixel (scaled from 0 to 1), and $t'$ is the time of the last spike.
    Although neurons with \ac{IF} neuron dynamics similar to the prediction and hidden neurons could be employed here, we assumed that data will be provided by external sensors in the form of spike trains that do not necessarily follow \ac{IF} dynamics.
    \reffig{fig:trace} illustrates the neural dynamics in a prediction neuron, in a network trained with 500 training samples (1/100 of an epoch).

    \subsubsection{Stochastic, blank-out Synapses and peRBP}
    In practice, we find that neurons tend to strongly synchronize in late stages of the training.  
    The analysis provided above does not accurately describe synchronized dynamics, since one of the assumptions for the diffusion approximation is that spike times are uncorrelated.
    Multiplicative stochasticity was previously shown to be beneficial for regularization and decorrelation of spike trains, while being easy to implement in neuromorphic hardware \cite{Neftci_etal16}.
    Following the ideas of synaptic sampling \cite{Neftci_etal16}, we find that replacing the background Poisson noise with multiplicative, blank-out noise \cite{Vogelstein_etal02} at the plastic synapses slightly improves the results and mitigates the energetic footprint of the stochasticity \cite{Sheik_etal16}.

    \subsection{Quantized, Discrete-time dynamics}
    In order to demonstrate the effectiveness of eRBP in dedicated digital hardware with realistic constraints on precision and limited computations, we created a simulation of a digital neuromorphic learning core.
    The hardware simulator is a bit-accurate, fixed point emulation of a spiking neural network implemented on a digital hardware platform for ultra-efficient and flexible learning dynamics that is currently under development. 
    Full details of the model and its architecture will be discussed elsewhere. 
    Here, we describe the subset of the model dynamics necessary for peRBP.
    The dynamics of the neuron $i$ implemented at the hidden layer is the following:
    \begin{equation}
      \begin{split}
        V^h_i [t+1] &= V^h_i[t] - m( a_V \diamond V^h_i[t] , V^h_i[t]) + a_{IV} \diamond I_i + b_V  \\
        U^h_i [t+1] &= U^h_i[t] - m( a_U \diamond U^h_i[t] , U^h_i[t]) +  \sum_j g_U \diamond w^E_{ij}[t] s^E_j[t]   \\
        I^h_i [t+1] &= I^h_i[t] - m( a_{syn} \diamond I^h_i[t], I^h_i[t] ) + \sum_j g_I \diamond w^h_{ij}[t] \xi[t] s^h_j[t] + \sum_j g_I \diamond w^d_{ij}[t] \xi[t] s^d_j[t] \\
    \text{if } V^h_i[t+1] &\ge V_T , s^h_i[t+1] \leftarrow 1 \text{ and } V^h_i[t+1] \leftarrow V_{reset} \\      
        \text{if } V^h_i[t+1] &< V_T   , s^h_i[t+1] \leftarrow 0 \\      
      \end{split}
    \end{equation}
    where the fourth and fifth line account for thresholds, resets, and spiking outputs $s_i$.
$s^d_j[t]$, $s^h_j[t]$, $s^h_j[t]$, and $s^E_j[t] \in \{ 0, 1 \}$ are the spiking output of the input (data), hidden, and error-coding neurons $i$ at time $t$, respectively.
    Indices from $\xi[t]$ were dropped for clarity, and every instance of $\xi[t]$ in the equations above refers to an independent and identically distributed Bernouilli draw with probability $p$.
    The terms $g_U$ and $g_I$ are weight gain factors used to adjust the range of the synaptic weights.

    Parameters $a$, including  $a_V$  $a_U$ $a_{syn}$, and $a_{IV}$ are integers implementing the coupling between and within the states $V$, $U$ and $I$.  
    The $\diamond$ operator is a custom bit shift that performs multiplication by powers of two and that can be implemented using only bitwise operations:    
    \begin{algorithmic}
      \Function{$a \diamond x$}{} 
    \If {$ a \ge 0 $}
    \textbf{return} $x\ll a$,
    \ElsIf {$ a < 0 $}
    \textbf{return} $sign(x)(|x| \gg -a$),
    \EndIf
    \EndFunction
    \end{algorithmic}
    The reason for using $\diamond$ rather than left and right bit shifting is because integers stored using a two's complement representation have the property that right shifting by $a$ of values such that $x>-2^{a'}, \forall a'<a$ is $-1$, whereas $0$ is expected in the case of a multiplication by $2^{-a}$.  
    The $\diamond$ operator corrects this problem by modifying the bit shift operation such that $-2^{a'}\diamond{a}=0,\,\forall a'<a$. 
    Furthermore, such multiplications by powers of $2$ have the advantage that fewer bits are required to \emph{store} parameters on a logarithmic scale, which is a natural parametrization for such linear difference equations.
    A similar operation was used in the BNN \cite{Courbariaux_Bengio16} for an approximate power-of-two operation, although in our simulations, the first argument $a$ is considered constant.

    The function $m(\cdot,\cdot)$ defined as
    \begin{algorithmic}
      \Function{$m(x, y)$}{} 
      \If {$ y \ne 0 $ and $x=0$}
      \textbf{return} $sign(-x)$,
    \Else
    \textbf{ return} $x$,
    \EndIf
    \EndFunction
    \end{algorithmic}
    ensures that all states leak towards zero in the absence of external drive.

    Weight dynamics are implemented as:
    \begin{equation}
      \begin{split}
          w^h_{ij} [t+1] &= \text{Clip}(w^h_{ij}[t] + \eta \diamond U^h_i[t] \Theta(I^h_i[t]) s_j[t], -128, 128),\\
      \end{split}
    \end{equation}
    where $\text{Clip}$ clips the weights higher than 128 and lower than -128.
    Dynamics for the prediction neurons were the same except that they reflected the connectivity of the output layer.
    Positive error neurons followed the following dynamics:
    \begin{equation}
      \begin{split}
        V^E_i [t+1] &= V^E_i[t] - m(a_V \diamond V^E_i[t], V^E_i) + (g_E \diamond w^{L}) (s_i^P(t) - s_i^L(t))   \\
    \text{if } V^E_i[t+1] &\ge V^E_T   , s^E_i[t+1] \leftarrow 1 \text{ and } V^E_i[t+1] \leftarrow V^E_i[t+1] - V_T^E \\      
    \text{if } V^E_i[t+1] &< V_T^E   , s^E_i[t+1] \leftarrow 0, \\      
      \end{split}
    \end{equation}
    As for continuous dynamics, negative error neurons follow the exact same dynamics with $w^{L}$ of opposite sign and error neuron membrane voltages are lower bounded to zero. 

    For data neurons, input spike trains were generated as Poisson spike trains with rate $\gamma d$, where $d$ is the pixel intensity. 
    For label neurons, input spikes were regular, \emph{i.e.} spikes were spaced regularly with interspike interval $\tau_{refr}^{-1}$

    In the simulations used for eRBP, all states $V$, $U$ and $I$ and parameters were stored in 16 bit fixed point precision (ranging from -32768 to 32767), except for synaptic weights which were stored with 8 bit precision (ranging from -128 to 128) and coupling parameters were stored with 5 bits precision (from 0 to 32).

    \subsection{Experimental Setup and Software Simulations}
    We trained fully connected feed-forward networks MNIST hand-written digits, separated in three groups, training, validation, and testing (50000, 10000, 10000 samples respectively).
    During a training epoch, each of the training digits were presented in sequence during $250\mathrm{ms}$.
    We tested eRBP using two configurations: one with additive noise ($\sigma_w>0$, $p=1$, labeled eRBP), and one with multiplicative noise implemented as blank-out noise on the connections (blank-out probability $p<1$ and $\sigma_w=0$, labeled peRBP).
    To prevent the network from learning (spurious) transitions between digits, the synaptic weights did not update in the first $50\mathrm{ms}$ window of each digit presentation.\\
    We tested eRBP training on two different implementations: 1) Spiking neural network based on the Auryn simulator \cite{Zenke_Gerstner14}, and 2) Hardware compatible simulator with quantized neural states and weights. Results are compared against GPU implementations of RBP and standard BP performed in Theano \cite{Bergstra_etal10} using an equivalent, non-spiking neural network. 
    Besides layered connectivity, all networks were unstructured (\emph{e.g.} no convolutions or poolings).\\

\begin{table*}
\begin{center}
    \begin{tabular}{|l | l | c | c |}
    \hline
      $N_d$           & Number of data  neurons                     & all networks                        & $784$\\
      $N_h$           & Number of hidden neurons                    &                                     & Variable\\
      $N_l$           & Number of label  neurons                    & all networks                        & $10$\\
      $N_{E+}$           & Number of positive error neurons         & all networks                        & $10$\\
      $N_{E-}$           & Number of negative error neurons         & all networks                        & $10$\\
      $N_p$           & Number of prediction neurons                & all networks                        & $10$\\
    $\sigma$        & Poisson noise weight                          & eRBP                                & $\unit[50\cdot10^{-3}]{nA}$ \\
                    &                                               & peRBP                               & $\unit[0\cdot10^{-3}]{nA}$ \\
    $p$             & Blank-out probability                         & eRBP                                & $\unit[1.0]{}$ \\
                    &                                               & peRBP                               & $\unit[.65]{}$ \\
    $\tau_{refr}$   & Refractory period                             & Prediction and hidden neurons       & $\unit[3.9]{ms}$\\
                    &                                               & Data neurons                        & $\unit[4.0]{ms}$\\
    $\tau_{syn}$    & Synaptic Time Constant                        & all synapses                        & $\unit[4]{ms}$\\
    $g_{V}$         & Leak conductance state $V$                    & Prediction and hidden neurons       & $\unit[1]{nS}$ \\
    $g_{U}$         & Leak conductance state $U$                    & Prediction and hidden neurons       & $\unit[5]{nS}$ \\
    $C$             & Membrane capacitance                          & all neurons                         & $\unit[1]{pF}$ \\
    $V_T   $        & Firing threshold                              & Prediction and Hidden neurons       & $\unit[100]{mV}$\\
    $V_T^E $        &                                               & Error neurons                       & $\unit[100]{mV}$\\
    $N_{train}$     & Number of training samples used               & all figures                         & $\unit[50000]{}$ \\
    $N_{test}$      & Number of training samples used               & \reftab{tab:erbp} eRBP, peRBP       & $\unit[10000]{}$\\
                    &                                               & \reftab{fig:edp} eRBP, peRBP        & $\unit[1000]{}$\\
                    &                                               & \reftab{fig:edp} RBP, BP            & $\unit[10000]{}$\\
    $T_{train}$     & Training sample duration                      & all models                          & $\unit[100]{mV}$ \\
    $T_{test}$      & Testing  sample duration                      & \reftab{tab:erbp},\reffig{fig:edp-spike}          & $\unit[500]{ms}$\\
                    &                                               & \reftab{fig:edp}                    & $\unit[250]{ms}$\\
    $w^h,w^d,w^p,g$ & Initial weight matrix                         & RBP, BP                             & $U(\sqrt{\frac{6}{\#rows+\#cols}})$\\
    $ $             &                                               & eRBP                                & $U(\sqrt{\frac{6}{\#rows+\#cols}})$nA\\
    $ $             &                                               &       peRBP                         & $U(\sqrt{\frac{7}{\#rows+\#cols}})$nA\\
    $w^{E }$        &                                               & eRBP, peRBP                         & $90\cdot10^{-3}$nA\\
    $w^{L+}$        &                                               & eRBP, peRBP                         & $90\cdot10^{-3}$nA\\
    $w^{L-}$        &                                               & eRBP, peRBP                         & $-90\cdot10^{-3}$nA\\
    $b_{min}$       &                                               & eRBP, peRBP                         & $\unit[-1.15]{nA}$\\
    $b_{max}$       &                                               & eRBP, peRBP                         & $\unit[ 1.15]{nA}$\\
    $\beta  $       & Data neuron input scale                       & eRBP, peRBP                         & $.5$\\
    $\gamma $       & Data neuron input threshold                   & eRBP, peRBP                         & $-.215$\\
    $T_{sim}$       & Simulation time per epoch                     & all models                          & $\unit[100]{s}$ \\
    $\epsilon$      & Learning Rate                                 & eRBP                                & $6\cdot10^{-4}$ \\
                    &                                               & peRBP                               & $10\cdot10^{-4}$ \\
                    &                                               & RBP, BP 1 hidden layer              & $.4$ \\
                    &                                               & RBP, BP 2 hidden layers             & $.5$ \\
    $n_{batch}$     & Batch size                                    & RBP, BP                             & $100$\\
    \hline
  \end{tabular}
\end{center}
\caption{\label{tab:parameters} Parameters used for the continuous-time spiking neural network simulation implementing eRBP.}
\end{table*}

\begin{table*}
\begin{center}
    \begin{tabular}{|l | l | c | c |}
    \hline
    $a_V,a_U,a_{syn},a_{IV}$ & State Couplings                                      & Hidden and prediction neurons       & $-3,-7,-6,4$ \\
    $g_V,g_U,g_I,g_E$& Synaptic weight gain factors                                 &                                     & $-3,-7,-6,4$ \\
    $V_T  $         & Firing threshold                                              & Prediction and hidden neurons       & $32797$ \\
    $V_T^E$         & Firing threshold                                              & Error neurons                       & $1025$\\
    $b_V$           & Bias                                                          & Prediction and hidden neurons       & $1000$\\
    $V_{reset}$     & Reset Voltage                                                 & Prediction and hidden neurons       & $32796$\\
    $p$             & Blank-out probability                                         & all plastic synapses                & $\unit[.6]{}$ \\
    $\eta$          & Learning rate                                                 & all neurons                         & $-10$ \\
    $\tau_{refr}$   & Refractory period                                             & Prediction and hidden neurons       & $\unit[39]{}$\\
                    &                                                               & Data neurons                        & $\unit[40]{}$\\
    $T_{sim}$       & Simulation time per epoch                                     & all models                          & $\unit[90\cdot10^6]{}$ \\
    $T_{train}$     & Training sample duration                                      & all models                          & $\unit[1500]{}$\\
    $T_{test}$      & Testing sample duration                                       & all models                          & $\unit[3000]{}$\\
    $b_{min}$       & eRBP lower bound                                              & all plastic synapses                & $\unit[-2560]{}$\\
    $b_{max}$       & eRBP upper bound                                              & all plastic synapses                & $\unit[ 2560]{}$\\
    $\beta  $       & Data neuron input scale                                       & data neurons                        & $25$\\
    $n_{batch}$     & Batch size                                                    & All GPU (Theano) simulations        & $100$\\
    \hline
  \end{tabular}
\end{center}
\caption{\label{tab:parameters} Parameters used for the discrete-time spiking neural network simulation implementing eRBP.}
\end{table*}

\section*{Acknowledgments}
This work was partly supported by the Intel Corporation and by the National Science Foundation under grant 1640081, and the Nanoelectronics Research Corporation (NERC), a wholly-owned subsidiary of the Semiconductor Research Corporation (SRC), through Extremely Energy Efficient Collective Electronics (EXCEL), an SRC-NRI Nanoelectronics Research Initiative under Research Task ID 2698.003.
We thank Jun-Haeng Lee and Peter O'Connor for review and comments; and Gert Cauwenberghs, Jo\~ao Sacramento, Walter Senn for discussion.

\bibliographystyle{plainnat}

\end{document}